\pdfoutput=1
\documentclass[sigconf, table, xcdraw]{acmart}
\AtBeginDocument{%
  \providecommand\BibTeX{{%
    \normalfont B\kern-0.5em{\scshape i\kern-0.25em b}\kern-0.8em\TeX}}}

\acmConference[Preprint]{Preprint}{Dec 2023}
\usepackage{tabularx}
\usepackage{listings,multicol}
\usepackage[shortlabels]{enumitem}
\usepackage{subcaption}
\usepackage{array}
\usepackage{multirow}
\usepackage{tikz}
\usepackage{amsmath}
\usepackage{amsthm}
\usepackage[export]{adjustbox}
\usepackage{booktabs}

\definecolor{codecomments}{rgb}{0,0.6,0}
\definecolor{codenumbers}{rgb}{0.5,0.5,0.5}
\definecolor{codestrings}{rgb}{.7, 0, .7}
\definecolor{codekeywords}{rgb}{0.8, 0.3, 0.4}

\lstdefinestyle{mystyle}{
    commentstyle=\color{codecomments},
    keywordstyle=\color{codekeywords},
    numberstyle=\tiny\color{codenumbers},
    stringstyle=\color{codestrings},
    basicstyle=\small\ttfamily,
    columns=flexible,
    breakatwhitespace=false,
    captionpos=b,
    keepspaces=true,
    numbers=left,
    numbersep=-4pt,
    showspaces=false,
    showstringspaces=false,
    showtabs=false,
    tabsize=2,
    frame=single
}

\begin{document}

\title{Pyreal: A Framework for Interpretable ML Explanations}

\author{Alexandra Zytek}
\email{zyteka@mit.edu}
\affiliation{%
  \institution{MIT}
  \city{Cambridge, MA}
  \country{USA}
}
\author{Wei-En Wang}
\email{weinwang@mit.edu}
\affiliation{%
  \institution{MIT}
  \city{Cambridge, MA}
  \country{USA}
}
\author{Dongyu Liu}
\email{dyuliu@ucdavis.edu}
\affiliation{%
  \institution{University of California, Davis}
  \city{Davis, CA}
  \country{USA}
}
\author{Laure Berti-{É}quille}
\email{laure.berti@ird.fr}
\affiliation{%
  \institution{IRD, ESPACE-DEV}
  \country{France}
}
\author{Kalyan Veeramachaneni}
\email{kalyanv@mit.edu}
\affiliation{%
  \institution{MIT}
  \city{Cambridge, MA}
  \country{USA}
}

\renewcommand{\shortauthors}{Zytek, et al.}

\begin{abstract}
Users in many domains use machine learning (ML) predictions to help them make decisions. Effective ML-based decision-making often requires explanations of ML models and their predictions. While there are many algorithms that explain models, generating explanations in a format that is comprehensible and useful to decision-makers is a nontrivial task that can require extensive development overhead. We developed \textsc{Pyreal}, a highly extensible system with a corresponding Python implementation for generating a variety of interpretable ML explanations. Pyreal converts data and explanations between the feature spaces expected by the model, relevant explanation algorithms, and human users, allowing users to generate interpretable explanations in a low-code manner. Our studies demonstrate that Pyreal generates more useful explanations than existing systems while remaining both easy-to-use and efficient.
\end{abstract}

\begin{CCSXML}
<ccs2012>
   <concept>
       <concept_id>10011007.10011006.10011072</concept_id>
       <concept_desc>Software and its engineering~Software libraries and repositories</concept_desc>
       <concept_significance>300</concept_significance>
       </concept>
   <concept>
       <concept_id>10010147.10010257</concept_id>
       <concept_desc>Computing methodologies~Machine learning</concept_desc>
       <concept_significance>500</concept_significance>
       </concept>
   <concept>
       <concept_id>10003120.10003121.10003122.10003334</concept_id>
       <concept_desc>Human-centered computing~User studies</concept_desc>
       <concept_significance>300</concept_significance>
       </concept>
 </ccs2012>
\end{CCSXML}

\ccsdesc[300]{Software and its engineering~Software libraries and repositories}
\ccsdesc[500]{Computing methodologies~Machine learning}
\ccsdesc[300]{Human-centered computing~User studies}

\keywords{machine learning, explainable machine learning, software, python}

%  \includegraphics[width=.7\textwidth]{figures/california_housing_both_demo_no_axis.pdf}


\maketitle

\newcommand{\Pyreal}{Pyreal}
\newcommand{\ttt}[1]{\texttt{#1}}
\newcommand{\note}[1]{\textcolor{red}{#1}}
\newcolumntype{P}[1]{>{\centering\arraybackslash}p{#1}}

\newcommand{\decisionfocused}{decision-focused}
\newcommand{\modelfocused}{model-focused}

\newcommand{\algo}{algo}
\newcommand{\Algorithm}{Algorithm}

% white number in grey circles
\definecolor{workflow_anno}{rgb}{0.64, 0.29, 0.12}
% rgb(164,74,30)

\makeatletter
\newcommand*\ccircle[2][draw]{\tikz[baseline=-0.65ex]
% {\node[circle,#1,inner sep=0,outer sep=0] (char) {\vphantom{WAH1g}#2};}
{\node[circle,#1,inner sep=.10ex] (char) {#2};}}
\makeatother
\newcommand{\nnbox}[1]{\ccircle[text=white,fill=workflow_anno]{#1}}

\newcommand{\E}{\mathbb{E}}
\newcommand{\F}{\mathbb{F}}

\newcommand{\tasknopyreal}{\ttt{no\_pyreal}}
\newcommand{\tasknopyrealclean}{no-\Pyreal{}}
\newcommand{\taskpyreal}{\ttt{pyreal}}
\newcommand{\taskpyrealclean}{\Pyreal{}}
\newcommand{\taskpyrealformat}{\ttt{pyreal\_format}}
\newcommand{\tasknopyrealinter}{\ttt{no\_pyreal\_\allowbreak interpret}}

\newcommand{\interpretable}{interpretable}  %as in, explanations
\newcommand{\Interpretable}{Interpretable}

\newcommand{\interpretablefs}{interpretable}  %as in, feature space
\newcommand{\Interpretablefs}{Interpretable}

\newcommand{\original}{original}  %as in, feature space
\newcommand{\Original}{Original}

\newcommand{\modelready}{model-ready}  %as in, feature space
\newcommand{\Modelready}{Model-ready}

\newcommand{\algoready}{algorithm-ready}  %as in, feature space
\newcommand{\Algoready}{Algorithm-ready}

\newcommand{\system}{system}
\newcommand{\System}{System}
\newcommand{\systems}{systems}
\newcommand{\Systems}{Systems}

\newcommand{\receivers}{receivers}
\newcommand{\Receivers}{Receivers}
\newcommand{\receiver}{receiver}
\newcommand{\Receiver}{Receiver}

\newcommand{\users}{bridges}
\newcommand{\Users}{Bridges}
\newcommand{\user}{bridge}
\newcommand{\User}{Bridge}

\newcommand{\developers}{developers}
\newcommand{\Developers}{Developers}
\newcommand{\developer}{developer}
\newcommand{\Developer}{Developer}

\newcommand{\contributors}{contributors}
\newcommand{\Contributors}{Contributors}
\newcommand{\contributor}{contributor}
\newcommand{\Contributor}{Contributor}
\section{Introduction}

\begin{figure}[h!t]
     \centering
     \begin{subfigure}[b]{\linewidth}
         \centering
         \includegraphics[width=\linewidth]{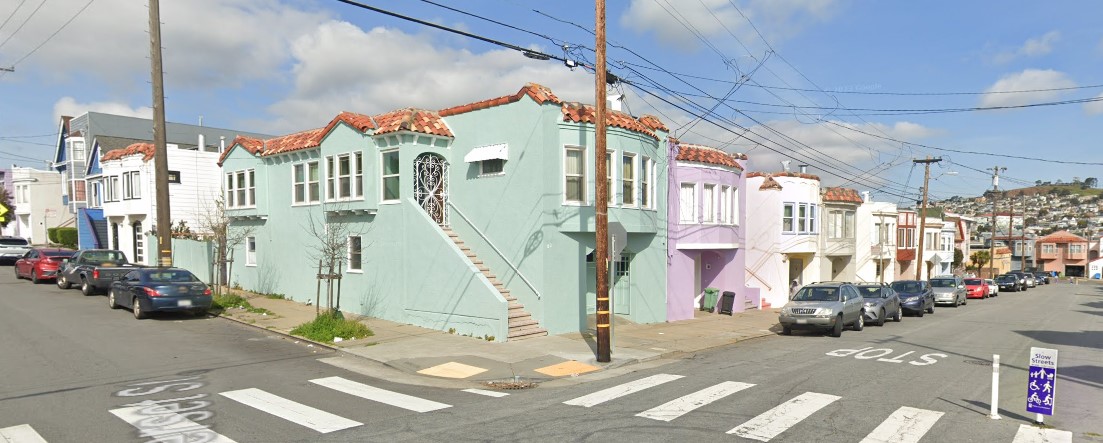}
         \caption{}

         \label{fig:sample_house} \vspace{1mm}
     \end{subfigure} 
     
     \begin{subfigure}[b]{\linewidth}
         \centering
         \includegraphics[width=\linewidth]{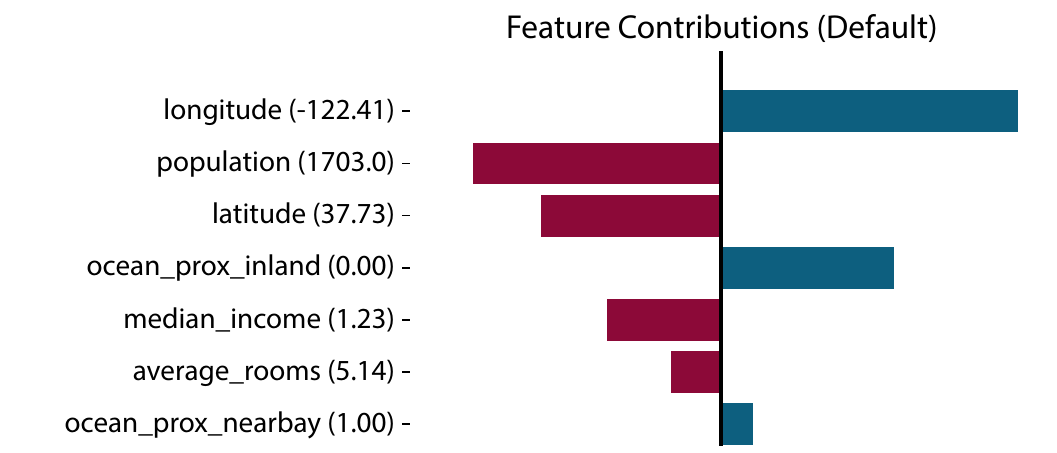}
         \caption{}
         \label{fig:cal_unint} \vspace{1mm}
     \end{subfigure} 
     
     \begin{subfigure}[b]{\linewidth}
         \centering
         \includegraphics[width=\linewidth]{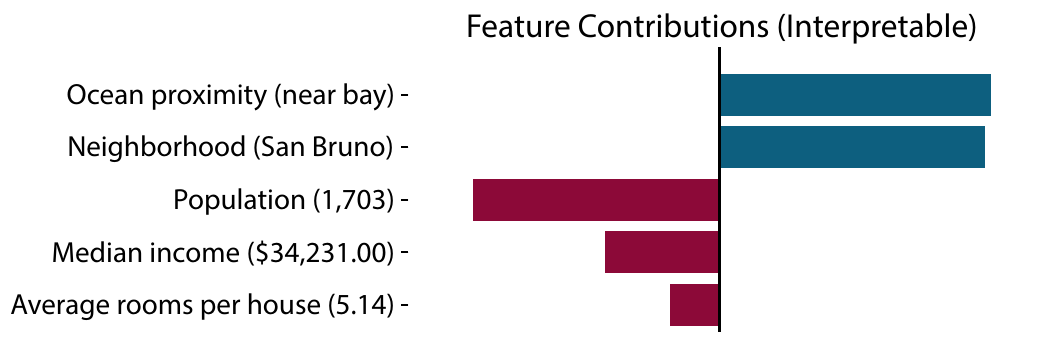}
         \caption{}
         \label{fig:cal_int}
     \end{subfigure} 
        \caption{Two ways of displaying an explanation of an ML model's prediction of the median house price of a block of houses like the one shown in (a), from the California Housing Dataset \cite{r_kelley_pace_sparse_1997}. 
        %of median house price for a block of houses within the California Housing Dataset \cite{r_kelley_pace_sparse_1997}. 
        The explanation in (b) shows the default output from existing explanation systems (\modelready{} feature space). The explanation in (c) shows a more readily-understandable explanation generated by \Pyreal{} (\interpretablefs{} feature space). Each bar represents how much a feature value increased (blue) or decreased (red) the model's prediction.  } \label{fig:motivation}

\end{figure}

% \usepackage{graphicx}
\begin{table*}[ht]
\caption{Comparison between \Pyreal{} and existing XAI libraries. \Pyreal{} is the only library to our knowledge that automatically generates interpretable explanations using a system of explanation transforms with a highly-extensible explanation API. The Explanation Type rows refer to the kinds of explanations that can be generated by the library. Model Type refers to the type of models that can be explained by the library. Flexible to ML library means models from more than one ML library can be explained (e.g., sklearn and lightgbm models). Agnostic to ML model type means any object with a standard model structure (i.e., a \ttt{.predict()} function) can be explained. OO refers to object-oriented classes.}
\label{tab:xai-libraries}
\resizebox{\textwidth}{!}{%
\begin{tabular}{lccccccccccccc}
\hline 
                                                            & \multicolumn{1}{c|}{}                                   & \multicolumn{2}{c|}{Algorithm Focused}                                                                                                                                                                                                               & \multicolumn{6}{c|}{Implements Multiple Explanation Algorithms}                                                                                                                                                                                                                                                                                                                                                                                                                                                      & \multicolumn{4}{c}{General ML Libraries that Include Explanation}                                                                                                                                                                                                                                                                                                     \\ \hline  
\multicolumn{1}{l|}{}                                       & \multicolumn{1}{c|}{\cellcolor[HTML]{EFEFEF}Pyreal}     & \begin{tabular}[c]{@{}c@{}}SHAP \\ \cite{lundberg_unified_2017}\end{tabular} & \multicolumn{1}{c|}{\begin{tabular}[c]{@{}c@{}}LIME \\ \cite{ribeiro_why_2016}\end{tabular}} & \begin{tabular}[c]{@{}c@{}}What-If Tool\\ \cite{wexler_what-if_2019}\end{tabular} & \begin{tabular}[c]{@{}c@{}}ELI5 \\ \cite{mikhail_korobov_eli5_2022}\end{tabular} & \begin{tabular}[c]{@{}c@{}}Alibi \\ \cite{JMLR:v22:21-0017}\end{tabular} & \begin{tabular}[c]{@{}c@{}}Skater \\ \cite{aaron_kramer_skater_2022}\end{tabular} & \begin{tabular}[c]{@{}c@{}}InterpretML \\ \cite{nori2019interpretml}\end{tabular} & \multicolumn{1}{c|}{\begin{tabular}[c]{@{}c@{}}AIX360 \\ \cite{arya_ai_2020}\end{tabular}} & \begin{tabular}[c]{@{}c@{}}Vertex AI \\ \cite{google_developers_vertex_nodate}\end{tabular} & \begin{tabular}[c]{@{}c@{}}scikit-learn \\ \cite{pedregosa_scikit-learn_2011}\end{tabular} & \begin{tabular}[c]{@{}c@{}}MLExtend \\ \cite{raschkas_2018_mlxtend}\end{tabular} & \begin{tabular}[c]{@{}c@{}}Yellowbrick \\ \cite{bengfort_yellowbrick_2019}\end{tabular} \\ \hline
\textbf{Explanation Type}                                   &                                                                                                                                       &                                                                         &                                                                                                    &                                                                                   &                                                                                  &                                                                          &                                                                                   &                                                                                   &                                                                                            &                                                                                             &                                                                                            &                                                                                  &                                                                                         \\ \hline
\multicolumn{1}{l|}{Multiple explanation Types}        & \multicolumn{1}{c|}{\cellcolor[HTML]{EFEFEF}\checkmark} & \checkmark                                                                                                                                                      & \multicolumn{1}{c|}{\checkmark}                                                                              & \checkmark                                                                        & \checkmark                                                                       & \checkmark                                                               & \checkmark                                                                        & \checkmark                                                                        & \multicolumn{1}{c|}{\checkmark}                                                            & \checkmark                                                                                  & \checkmark                                                                                 & \checkmark                                                                       & \checkmark                                                                              \\
\multicolumn{1}{l|}{Supports global and local explanations} & \multicolumn{1}{c|}{\cellcolor[HTML]{EFEFEF}\checkmark} & \checkmark                                                                   &                                                                          \multicolumn{1}{c|}{}                                                                              & \checkmark                                                                        & \checkmark                                                                       & \checkmark                                                               & \checkmark                                                                        & \checkmark                                                                        & \multicolumn{1}{c|}{\checkmark}                                                            & \checkmark                                                                                  &                                                                                            &                                                                                  &                                                                                         \\ \hline
\textbf{Use-case Flexibility}                                         &                                                         &                                                                              &                                                                         &                                                                                                    &                                                                                                                                                                     &                                                                          &                                                                                   &                                                                                   &                                                                                            &                                                                                             &                                                                                            &                                                                                  &                                                                                         \\ \hline
\multicolumn{1}{l|}{Supports classification and regression} & \multicolumn{1}{c|}{\cellcolor[HTML]{EFEFEF}\checkmark} & \checkmark                                                                   &                                                                \multicolumn{1}{c|}{\checkmark}                                                                              & \checkmark                                                                        & \checkmark                                                                       & \checkmark                                                               & \checkmark                                                                        & \checkmark                                                                        & \multicolumn{1}{c|}{\checkmark}                                                            & \checkmark                                                                                  & \checkmark                                                                                 & \checkmark                                                                       & \checkmark                                                                              \\
\multicolumn{1}{l|}{Agnostic to ML model type}              & \multicolumn{1}{c|}{\cellcolor[HTML]{EFEFEF}\checkmark} & \checkmark                                                                   &                                                               \multicolumn{1}{c|}{\checkmark}                                                                    & \checkmark                                                                        & \checkmark                                                                       & \checkmark                                                               & \checkmark                                                                        & \checkmark                                                                        & \multicolumn{1}{c|}{\checkmark}                                                            & \checkmark                                                                                  &                                                                                            &                                                                                  &     -                                                                                    \\ \hline
\textbf{Library Stability}                                   &                                                         &                                                                              &                                                                         &                                                                                                                                                                                       &                                                                                  &                                                                          &                                                                                   &                                                                                   &                                                                                            &                                                                                             &                                                                                            &                                                                                  &                                                                                         \\ \hline
\multicolumn{1}{l|}{Benchmarking/Evaluation}                & \multicolumn{1}{c|}{\cellcolor[HTML]{EFEFEF}\checkmark} & \checkmark                                                                   &                                                                \multicolumn{1}{c|}{\checkmark}                                                                              &                                                                                   &                                                                                  &                                                                          &                                                                                   & \checkmark                                                                        & \multicolumn{1}{c|}{\checkmark}                                                            &                                                                                             &                                                                                            &                                                                                  &                                                                                         \\
\multicolumn{1}{l|}{Open source implementation}             & \multicolumn{1}{c|}{\cellcolor[HTML]{EFEFEF}\checkmark} & \checkmark                                                                   &                                                              \multicolumn{1}{c|}{\checkmark}                                                                    & \checkmark                                                                        & \checkmark                                                                       & \checkmark                                                               & \checkmark                                                                        & \checkmark                                                                        & \multicolumn{1}{c|}{\checkmark}                                                            &                                                                                             &                                                                                            & \checkmark                                                                       & \checkmark                                                                              \\ \hline
\textbf{Ease of Switching Explanations}                             &                                                         &                                                                                                                                                       &                                                                                                    &                                                                                   &                                                                                  &                                                                          &                                                                                   &                                                                                   &                                                                                            &                                                                                             &                                                                                            &                                                                                  &                                                                                         \\ \hline
\multicolumn{1}{l|}{Consistent Explainer API}         & \multicolumn{1}{c|}{\cellcolor[HTML]{EFEFEF}\checkmark} &                                                                              &                                                                          \multicolumn{1}{c|}{}                                                                              &                                                                                   & \checkmark                                                                       & \checkmark                                                               &                                                                                   &                                                                                   & \multicolumn{1}{c|}{\checkmark}                                                            &                                                                                             &                                                                                            &                                                                                  &                                                                                         \\
\multicolumn{1}{l|}{Consistent Explanation Type API}       & \multicolumn{1}{c|}{\cellcolor[HTML]{EFEFEF}\checkmark} &                                                                              &                                                                          \multicolumn{1}{c|}{}                                                                              &                                                                                   &                                                                                  &                                                                          &                                                                                   &                                                                                   & \multicolumn{1}{c|}{}                                                                      &                                                                                             &                                                                                            &                                                                                  &                                                                                         \\
\hline
\textbf{Human-Centered Approach}                            &                                                                                                                                       &                                                                         &                                                                                                    &                                                                                   &                                                                                  &                                                                          &                                                                                   &                                                                                   &                                                                                            &                                                                                             &                                                                                            &                                                                                  &                                                                                         \\ \hline
\multicolumn{1}{l|}{Explanations Targeted to decision makers}           & \multicolumn{1}{c|}{\cellcolor[HTML]{EFEFEF}\checkmark} &                                                                              &                                                                          \multicolumn{1}{c|}{}                                                                              & \checkmark                                                                        &                                                                                  &                                                                          &                                                                                   &                                                                                   & \multicolumn{1}{c|}{\checkmark}                                                            & \checkmark                                                                                  &                                                                                            &                                                                                  &                                                                                         \\
\multicolumn{1}{l|}{Explanation transforms}                 & \multicolumn{1}{c|}{\cellcolor[HTML]{EFEFEF}\checkmark} &                                                                              &                                                                          \multicolumn{1}{c|}{}                                                                              &                                                                                   &                                                                                  &                                                                          &                                                                                   &                                                                                   & \multicolumn{1}{c|}{}                                                                      &                                                                                             &                                                                                            &                                                                                  &                                                                                         \\ \hline
\end{tabular}%
}
\end{table*}

Machine learning has the potential to drive a productivity revolution. If effectively put in the hands of end users, ML tools could enable efficient and accurate problem solving across a diverse range of domains. However, a lack of truly useful explanatory ML systems is preventing this leap. 

To illustrate, we present a hypothetical: A real estate agent is helping their client put in a winning offer for a house. In their decision-making, the agent uses an ML model trained to predict the median price of houses on a block.

To better understand the factors that contribute to median house price prediction for a certain block of houses in California, the agent looks at the explanation shown in Figure \ref{fig:cal_unint} (a feature contribution explanation). They will likely struggle to make sense of what they see there: for instance, how latitude (-122.41) contributes negatively, or longitude (37.73) contributes positively. Additionally, what do \ttt{ocean\_prox\_inland} (0.00) and \ttt{ocean\_prox\_nearbay} (1.00) mean, and how can a median income be 1.23? 

Now, imagine the same agent looking at the same ML model prediction via the explanation shown in Figure \ref{fig:cal_int}. The agent now understands that the block is near the bay and in San Bruno, and its location has contributed to a higher price. Meanwhile, the median income for the block is now a meaningful \$34,231.
%This type of interaction makes the ML model usable. 

Most current systems for generating explanations of ML models and predictions give outputs like our first example. ML model developers, in a quest to improve predictive accuracy, transform data multiple times before using it to train an ML model. Explanation methods are usually model-focused, and therefore return explanations that use this (often uninterpretable) transformed data.

To address this problem, we leverage the \textit{\interpretablefs{} feature space} \cite{zytek_need_2022}, or the format of the data most easily understood by end users. Explanations presented using the \interpretablefs{} feature space --- such as Figure \ref{fig:cal_int} --- are called \textit{\interpretable{} explanations}. 
When the feature space used by the model is not inherently \interpretable{}, \interpretable{} explanations can be generated using \textit{explanation transforms}, which transform an existing explanation to an equivalent explanation in the \interpretablefs{} feature space. 

Generating \interpretable{} explanations in this way poses a significant systems problem. While often not theoretically complicated, transforming explanations to interpretable states often 
%for many common explanation types and transforms, 
requires extensive overhead code. Each time a new explanation algorithm is tried, or a new data transform is introduced, new code is needed. Especially since real-world deployments often involve experimenting with many different explanations \cite{zytek_sibyl_2021}, this code requirement is impractical and introduces opportunities for errors. 

The systems problem is made more difficult by the immense number of data transforms and explanation algorithms that exist, with more being developed all the time. Therefore, for an \interpretable{} explanation system to remain effective, it must be flexible to a diverse range of problems, and structured so as to allow for the integration of new explanation methods far into the future. 

\begin{figure*}
    \centering
    \includegraphics[width=\linewidth]{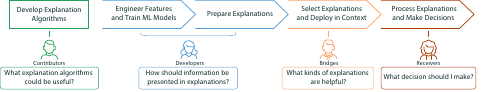}
    \caption{The ML explanation workflow, and the kinds of people involved. The \Pyreal{} \system{} is built with all these roles in mind.} 
    \label{fig:roles}
\end{figure*}

In this paper, we introduce \Pyreal{}\footnote{https://github.com/sibyl-dev/pyreal}, a highly extensible system for generating \interpretable{} ML explanations for any ML use case with tabular data. This work offers the following contributions:

\setlist{nolistsep}
\begin{itemize}[noitemsep]
    \item \textbf{\Interpretable{} Explanations. } To our knowledge, \Pyreal{} is the first system that takes data in any feature space and automatically returns a variety of \interpretable{} explanations using explanation transforms.
    \item \textbf{User-Aware API. } \Pyreal{} requires minimal Python knowledge, and can create useful explanations with just a few simple lines of code.
    \item \textbf{Explanation Substitution.} The \Pyreal{} API allows users to switch between different types of explanations by changing a single line of code.
    \item \textbf{Highly-Extensible and Flexible System.} We follow the example of existing ML \systems{} such as \ttt{sklearn} \cite{pedregosa_scikit-learn_2011} in developing a modular \system{} that sets up a common API to enable easy addition of new explanation and transformation components. The \system{} can support any model or transformer with a standard \ttt{.predict()} or \ttt{.transform()} function, respectively, that runs on tabular data.
    We focus on tabular data, as it is very common in real-world decision problems.
\end{itemize}

Our illustrative examples in this paper use the \textbf{California Housing Dataset} \cite{r_kelley_pace_sparse_1997}. This dataset has nine features, and each row describes a block of houses in California. The target variable our model predicts is the median price of housing for each block. 

\section{Related Work} \label{sec:related_work}
Much work has been done on the most effective ways to explain ML predictions to decision makers. This is an essential part of the \textit{human-centered ML perspective} \cite{gillies_human-centred_2016}. 

Explanations can be \textit{global}, meaning they explain the model's general prediction process \cite{doshi-velez_towards_2017}, or \textit{local}, meaning they explain a model's prediction on a specific input \cite{doshi-velez_towards_2017}.

Zhou et al. \cite{zhou_evaluating_2021} characterize a good explanation as having fidelity and being interpretable. Explanations with fidelity accurately describe the model, while interpretable explanations are unambiguous and presented simply. Our system aims to improve the interpretability of existing, fidelitous explanation algorithms. 
Explanations can be made more interpretable by
ensuring a reasonable cognitive load \cite{abdul_cogam_2020, cheng_explaining_2019, colin_what_2023}, utilizing positive framing \cite{hadash_improving_2022}, and avoiding the use of confounding ML transformations \cite{jiang_two_2021, zytek_need_2022}. \Pyreal{} focuses primarily on this last point.

Several authors have addressed the need to consider individuals with different types of expertise when working with explainable ML applications \cite{barredo_arrieta_explainable_2020, bhatt_explainable_2020, langer_what_2021, preece_stakeholders_2018, zytek_lessons_2023}. Individuals working with or affected by ML explanations may be ML experts, but they also may have little to no ML or coding experience. 
We aim to consider the full range of individuals that interact with our system or its explanations.

Many \systems{} already exist for generating ML explanations. Table \ref{tab:xai-libraries} compares the capabilities of \Pyreal{} to other \systems{} that offer multiple types of ML explanations. \Pyreal{} matches or exceeds other \systems{} on several axes. It offers both global and local explanation options. It is flexible to use-case, as it can explain any model accepting a tabular feature matrix. Additionally, the \Pyreal{} \system{} offers stability and developer-friendliness through a benchmarking approach that ensures the code remains error-free, and by having an open-source implementation.

\Pyreal{} is one of the few ML explanation \systems{} that allows for fast and easy switching of explanation algorithms, allowing users to change only a method call without further changes to overhead code. To our knowledge, it is also the only such \system{} that has a unified structure for explanation types, allowing for explanations generated using different algorithms to be switched out seamlessly, as well as the only one that takes an explicitly human-centered approach by generating \interpretable{} explanations through a pipeline for transforming explanations. 

\section{Roles in the \Interpretable{} Explanation Workflow}

We consider four roles people may fill when interacting with ML explanations or explanation systems such as \Pyreal{}, summarized in Figure \ref{fig:roles}.
These are \textbf{\receivers{}}, who receive interpretable ML explanations and use them for their decision-making; \textbf{\users{}} \cite{zytek_lessons_2023}, who help bridge the gap between receivers and ML experts by selecting and tuning appropriate explanations; \textbf{\developers{}}, who train ML models and prepare explanations, and \textbf{\contributors{}}, who expand explanation \systems{}.

To put these four roles in context, consider this hypothetical scenario (which matches the process we have observed in real-world domains): An ML engineer (\contributor{}) sees value in feature importance explanations computed using LIME \cite{ribeiro_why_2016}, so she adds this algorithm to \Pyreal{}. Another ML engineer (\developer{}), working for a firm that offers tools for the real estate industry, trains a model that predicts house prices. He then prepares an application that enables easy generation of explanations for the model,
%(through a process explained later)
and sends it over to an interested real estate agent. This agent (\user{}/\receiver{}) uses the application to get model predictions and explanations to help her decide what opening offer price to suggest to her clients. She then selects explanations she believes will be helpful to show to her clients (\receivers{}) to motivate her suggested offer.

We will now dive into how each of these roles interacts with the \Pyreal{} \system{}, and in doing so discuss the \system{}'s API and underlying structure.
%We will now dive into each of these roles in further detail, and in doing so discuss the full \Pyreal{} system. 
\section{\Receivers{}: Effectively Understanding Explanations}
\Pyreal{} explanations are built specifically for a target audience of people who are not familiar with ML or data science, though they are often domain experts. Explaining ML models in a way that improves decision-making is a highly context-dependent task \cite{barredo_arrieta_explainable_2020, hase_evaluating_2020, hong_human_2020} that requires iterating within and understanding the specific domain \cite{zytek_sibyl_2021, jiang_two_2021, nyre-yu_considerations_2021}. In particular, which features are considered ``interpretable'' will depend on the expertise of decision-makers \cite{zytek_need_2022}. \Pyreal{} does not remove the need to understand the intricacies of an individual domain through active collaborations, but rather aids with the generation of explanations that meet identified interpretability needs.

We assume a target audience of \receivers{} who are not interested in how the model itself works (as an ML model developer may be), but rather what the model or explanations can tell them about the context of the decision-making process. For example, our \receivers{} are likely not interested in the fact that the model uses a normalized metric for median income, but would be interested to learn that the low median income in a neighborhood reduced a house price prediction. Transforming explanations to a more \interpretablefs{} feature space helps provide the latter type of information.

\begin{figure}[tb]
    \centering
    \begin{lstlisting}[language=Python]
   from pyreal import visualize

   # Load in premade realapp and data (external functions)
   app = load_realapp()
   X_to_explain = load_data_to_be_explained() 

   exp1 = app.produce_feature_contributions(X_to_explain)
   exp2 = app.produce_similar_examples(X_to_explain)

   visualize.feature_bar_plot(exp1)   \end{lstlisting}
   \caption{\User{} workflow. \Users{} use RealApp objects to produce a variety of explanations, each with a single line of code. Figure \ref{fig:cal_int} has a sample output of the final visualize function.} 
         \label{fig:pyreal_user_code}
\end{figure}

\section{\Users{}: Efficiently Generating Explanations} \label{sec:using_pyreal}

The term \textit{\users{}} refers broadly to the people who exist across domains who help bridge the gap between ML developers and receivers \cite{zytek_lessons_2023}. These individuals have enough knowledge of the problem at hand to identify what kind of information would be helpful for receivers, while also having enough understanding of ML to know what kinds of explanations exist. Their job in the ML explanation workflow is to collaborate with both developers and receivers to select useful explanations.

\Pyreal{}'s high-level API enables \users{} to generate a wide variety of explanations in a low-code manner. Figure \ref{fig:pyreal_user_code} shows the minimal code required to produce and visualize an explanation as a \user{}. 
%We note that as basic coding knowledge becomes more widespread, the receiver and \user{} role can increasingly be held by the same people.

\Pyreal{}'s high-level API is realized through RealApps --- fully encapsulated ML explanation applications that are prepared by developers, as discussed in Section \ref{sec:developing_pyreal}. 
%hold all Pyreal components needed to produce explanations.
%will prefit these RealApp objects (as discussed in the next section) such that they can automatically produce interpretable explanations of any kind currently supported by Pyreal.
\Users{} use RealApp objects primarily through their \textit{produce} functions, one of which exists per \textit{explanation type} --- in other words, the kind of information offered by the explanation. For example, the \textit{produce feature importance} function returns one positive value for each feature, representing that feature's overall importance, and the \textit{produce similar examples} function gives a few examples from the training dataset that are similar to the input data. \Users{} do not need to consider the specific algorithm used to generate the explanations (ex. SHAP, relative weights, etc. for feature importance), which can be selected by developers or chosen by \Pyreal{} based on the current state-of-the-art. We chose this design as \users{} are more concerned about what information they want to see rather than the specific algorithmic details used to get that information, which requires more extensive ML expertise.
%They hold a list of explainer objects (discussed in depth in Section \ref{sec:explainers}), each tuned to generate a specific type of explanation, which are fit on their first use or can be pre-fit. They also hold 

RealApp objects output explanations in data structures that make them easy to access and use. %, making them easy to access, visualize, and plug into external visualization and interaction tools. 
For example, feature contribution explanations are indexed by row IDs, and include a pandas DataFrame with the feature name, value, and contribution columns for each row. This data can then be plugged into and visualized with systems such as \Pyreal{}'s visualization module, Tableau, or our generalizable UI system\footnote{https://github.com/sibyl-dev/sibylapp2}.

\begin{figure}[tb]
    \centering
    \begin{lstlisting}[language=Python]
   from pyreal import RealApp
   from pyreal.transformers import (
      Imputer, OneHotEncoder, NeighborhoodEncoder)

   transformers = [Imputer(model=True, interpret=True),
                   OneHotEncoder(
                       columns="ocean_proximity", 
                       model=True, interpret=False),
                   LatLongToNeighborhoodEncoder(
                       model=False, interpret=True)]

   # Load in data and model (external functions)
   data = load_training_data() 
   model = train_model(data, transformers)  
                    
   feature_descs = {"med_income": "Median income",
                    "ocean_prox": "Ocean Proximity",
                    ...}

   app = RealApp(model, X_train_orig=data, 
                 transformers=transformers,
                 feature_descriptions=feature_descs)  \end{lstlisting}
   \caption{\Developer{} workflow. \textbf{Step 1 (lines 5-10):} the required \Pyreal{} transformers for the application are selected, including those that are required to run the model (feature engineering) and those that make data more \interpretable{}. The \ttt{model} and \ttt{interpret} flags are used to label them accordingly. In this example, imputing data and one-hot-encoding are required for the model, but reduce the interpretability of the data. Additionally, this model takes in a latitude and longitude representation of location, but a categorical feature referring to the neighborhood is more interpretable.
        \textbf{Step 2 (lines 12-13):} The data is loaded in and the ML model is trained, as in traditional ML workflows.
        \textbf{Step 3 (lines 15-21):} A RealApp object is initialized, providing a feature description dictionary to improve the readability of feature names. This RealApp object can now be used to produce \interpretable{} explanations, as in Figure \ref{fig:pyreal_user_code}.
        } \label{fig:pyreal_dev_code}
\end{figure}

\begin{figure*}[t]
    \centering 
    \includegraphics[width=.9\linewidth, right]{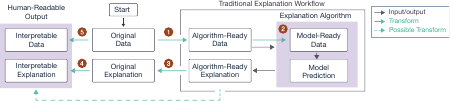}
    \caption{The \interpretable{} explanation transformer pipeline. Each step refers to either data or an explanation in one of the four \Pyreal{} feature spaces. The process starts at the ``start'' box, with the \original{} data. \nnbox{1} Data in the \original{} feature space is transformed to the \algoready{} feature space. \nnbox{2} The explanation algorithm runs on data in the \algoready{} feature space. Most explanation algorithms involve making predictions with the model, so data is also transformed into the \modelready{} feature space as part of this process. \nnbox{3} The explanation is converted to the \original{} feature space by ``undoing'' any transforms that reduced interpretability. \nnbox{4} The explanation is converted to the \interpretable{} feature space. In some cases it may be practical to convert explanations directly from the \algoready{} feature space to the \interpretable{} feature space, when these two spaces are very similar. \nnbox{5} For local explanations, the input that is being explained is transformed to the \interpretable{} feature space, so its values can be displayed alongside the explanation.
    }
    \label{fig:transformer_workflow}
\end{figure*}

\section{\Developers{}: Making \Interpretable{} Explanation Apps} \label{sec:developing_pyreal}

\Developers{} are ML experts who train models and prepare explanation applications for new domains. They have the expertise required to train ML models and perform feature engineering, and work with \users{} to gain the domain expertise necessary to understand what features are \interpretable{} to their \receivers{}. Within the \Pyreal{} workflow, \developers{} create RealApp objects and tune them for the decision-making problem at hand. 

All components needed to produce \interpretable{} ML explanations for a decision-making problem are encapsulated in and managed by a single RealApp object. These components include one or more ML models, objects that run explanation algorithms, objects that transform data and explanations, and other context-specific information. All these components are explained in Section \ref{sec:contributing_pyreal}.

%(these components are explained in depth in Section \ref{sec:contributing_pyreal}). 
\Developers{} only need to consider how data and explanations need to be transformed to fit their application (they can select transformers from the growing set already implemented in \Pyreal{}, or implement domain-specific ones as necessary), and handle the ML model preparation and training. The full \developer{} workflow is detailed in Figure \ref{fig:pyreal_dev_code}.

\section{\Contributors{}: Building an \Interpretable{} Explanation System} 
\label{sec:contributing_pyreal}

\Contributors{} are explainable ML experts who are responsible for expanding ML explanation \systems{} like \Pyreal{}.

The ML community is constantly developing new explanation algorithms and new transformers. Different domains may require very different transformers or explanations. 
Our single team cannot provide all possible components required for any domain. Therefore, \Pyreal{} is built to be highly extensible. 

This extensibility is provided in two ways. First, \Pyreal{} generalizes the transform pipeline necessary to generate interpretable explanations. As part of this pipeline, we define four \textit{feature spaces} into which data and explanations may need to be transformed to generate interpretable explanations.
Second, \Pyreal{} defines a modular and hierarchical class structure, so no functionality needs to be re-implemented. Base classes handle all the overhead code necessary to run the feature space pipeline. Explanations are categorized by output type, unifying algorithms that generate explanations that offer the same kind of information.

In this section, we discuss the generalized \Pyreal{} pipeline and modular components and summarize the contributor workflow involved in expanding the system.

\subsection{The Four Feature Spaces of Interpretable Explanations} \label{sec:feature_spaces}

A \textit{feature space} is defined as data in a specific format that serves a specific purpose. For example, a numeric feature may be in an imputed and standardized feature space. Feature spaces are transformed to other feature spaces using one or more transformers. The \Pyreal{} pipeline considers four feature spaces:

\begin{sloppypar}
\textbf{Original Feature Space.} The format in which the data originally exist.

\textbf{Model-Ready Feature Space.} The format of data that the model requires to make predictions. 

\textbf{\Algorithm-Ready Feature Space.} The format of data that the explanation algorithm expects.  

\textbf{Interpretable Feature Space.} The format of the data that is most useful or natural for users.
\end{sloppypar}

To better explain how these four feature spaces work in practice, consider the example of features that convey information about the location of a block of houses, such as in the California Housing Dataset. Also, consider the common situation of an ML model that requires all features be passed to it in a numeric format, and an explanation algorithm that determines feature importance through permuting feature values. This information could be presented as follows:
 
\textbf{Original:} Numeric latitude and longitude features \\  \ttt{lat} = 37.7, \ttt{long} = -122.5

\textbf{Model-ready:} Boolean one-hot encoded neighborhood \\ \ttt{neighborhood\_is\_oceanview = True}

\textbf{Algorithm-ready:} Categorical neighborhood feature \\  \ttt{neighborhood =} \ttt{Oceanview}

\textbf{Interpretable:} Categorical neighborhood with city label \\ \ttt{neighborhood = }\ttt{Oceanview (San Francisco)}

Note that the algorithm feature space is categorical, as permuting one-hot encoded features may result in impossible rows of data where multiple or zero one-hot encoded columns are set to True. The model space could be latitude/longitude values or one-hot-encoded neighborhood values, depending on which improves model performance.

In many cases, there is overlap between these spaces. For example, a transformer may produce a feature that is both predictive (model-ready) and useful to users (interpretable). Additionally, most explanation algorithms take data in the model-ready feature space. 
%In total, there are 10 unique ways that a feature may need to move through the four feature spaces.

Explanations can also exist in different feature spaces, as explanations present information in terms of features. Figure \ref{fig:motivation} shows an example of the same explanation in two feature spaces (model-ready and interpretable). Transformers therefore also need to operate on explanations. The method for converting a data transform to a corresponding explanation transform depends on the transform and explanation type. 

% \usepackage{multirow}
% \usepackage{graphicx}
\begin{table*}[t]
\centering
\caption{Sample data/inverse explanation transforms on two categories of explanations. Note that [inverse] explanation transformations are not always possible in all cases; Pyreal will automatically stop the transform process whenever necessary on both explanations and data.}
\label{tab:sample-transforms}
\resizebox{\textwidth}{!}{%
\begin{tabular}{
>{\columncolor[HTML]{EFEFEF}}l lll}
\hline
\cellcolor[HTML]{EFEFEF} &
  \multicolumn{1}{c}{} &
  \multicolumn{2}{c}{\textbf{Sample Inverse Explanation Transformations}} \\
\multirow{-2}{*}{\cellcolor[HTML]{EFEFEF}\textbf{Transfomer}} &
  \multicolumn{1}{c}{\multirow{-2}{*}{\textbf{Data Transformation}}} &
  \multicolumn{1}{c}{\textbf{\begin{tabular}[c]{@{}c@{}}Additive Feature Contributions/Importance\\ (SHAP)\end{tabular}}} &
  \multicolumn{1}{c}{\textbf{\begin{tabular}[c]{@{}c@{}}Example-Based Explanations\\ (Counterfactuals, Prototypes, etc.)\end{tabular}}} \\ \hline
\textbf{One-hot Encoder} &
  \begin{tabular}[c]{@{}l@{}}Converts categorical features to set of \\ boolean  features\end{tabular} &
  \begin{tabular}[c]{@{}l@{}}Sum together contributions of one-hot \\ encoded features\end{tabular} &
  \begin{tabular}[c]{@{}l@{}}Decode one-hot encoded feature back \\ to categorical\end{tabular} \\
\textbf{Standardizer/Normalizer} &
  \begin{tabular}[c]{@{}l@{}}Standardizes/normalizes numeric features \\ to have consistent mean or range\end{tabular} &
  No change needed &
  \begin{tabular}[c]{@{}l@{}}Undo normalization/standardization using \\ saved constants\end{tabular} \\
\textbf{Feature Selector} &
  \begin{tabular}[c]{@{}l@{}} Select a subset of features to feed into \\the model\end{tabular} &
  Set contribution/importance score to 0 &
  Set unused features to ``any'' \\
\textbf{Numeric binning} &
  \begin{tabular}[c]{@{}l@{}}Assign numeric values to categorical \\ bins based on values\end{tabular} &
  \begin{tabular}[c]{@{}l@{}}Take contribution from bin corresponding \\to original value\end{tabular} &
  Leave as bin (specify thresholds) \\
\textbf{Combined categories} &
  \begin{tabular}[c]{@{}l@{}}Combine categories for categorical \\ features into broader categories\end{tabular} &
  \begin{tabular}[c]{@{}l@{}}Take contribution from parent corresponding \\ to original child category\end{tabular} &
  Leave as bin (specify child categories) \\ \hline
\end{tabular}%
}
\end{table*}

\subsection{The Pyreal Feature Space Pipeline} \label{sec:process}
Most ML explanation systems (such as those introduced in Table \ref{tab:xai-libraries}) take data in the feature space expected by the model or explanation algorithm, and return explanations in the same model- or algorithm-ready feature space. 
However, in most real-world use cases, the explanation pipeline begins when the data first reach the code (in the original feature space) and the explanation needs to be presented to the receivers in an interpretable feature space\footnote{Note that some \receiver{} groups, like ML engineers looking to improve a model, may prefer model-ready explanations. The pipeline still supports these scenarios; the interpretable feature space simply becomes equal to the model-ready feature space}. We refer to this as an interpretable explanation pipeline.

The interpretable explanation pipeline used by \Pyreal{} is detailed in Figure \ref{fig:transformer_workflow}. 
Note that not every feature needs to go through the full transform process. For example, a transform that makes data both algorithm-ready and more interpretable does not need to be ``undone'' at step 3. 
Through this system of transforms, \Pyreal{} generates interpretable explanations from data, regardless of the input feature state. 

\subsection{The Extensible Pipeline Implementation}

\begin{figure}[tb]
    \centering
    \includegraphics[width=\linewidth]{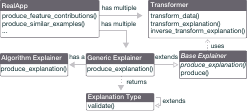}
    \caption{\Pyreal{} class structure. Italics refers to abstract classes or methods. The workflow of interactions between these classes is as follows:  RealApps contain all the necessary explainers and transformers for the interpretable explanation pipeline. Explainers use transformers to transform input and training data to the feature space they require --- to avoid code reuse, this process is handled by parent base explainer classes --- and then run explanation algorithms on this data. They return the explanation using the appropriate explanation type class from the type heirarchy. The explainer then uses the same transformers on this explanation type object to transform it to the interpretable feature space.}
    \label{fig:class-structure}
\end{figure}

\Pyreal{}'s internal implementation includes three class types: explainers, which run explanation algorithms; transformers, which transform data and explanations between feature spaces; and explanation types, which unify the transform process between explanations with similar properties.
Currently, we offer a starting set of six explainers, ten transformers, and ten explanation output types.

Modularity through these classes enables \Pyreal{} to stay up-to-date with state-of-the-art tabular explanation methods while remaining backward compatible, and prevents the need for major refactors or code reuse. 
%More details on the structure of \Pyreal{} can be found in our documentation\footnote{https://github.com/sibyl-dev/pyreal}.

The \Pyreal{} class structure is summarized in Figure \ref{fig:class-structure}, and discussed in further detail through the rest of this section. RealApp objects were introduced in earlier sections; we now discuss the remaining three class types.

\textbf{Explainer} objects generate interpretable explanations. They have a \ttt{fit} method that performs any computation-intensive procedures to prepare for generating explanations and a \ttt{produce} method that generates an explanation. For local explainers, \ttt{produce} takes in input data and generates an explanation for the model prediction on that input; for global explainers, the method takes no arguments and generates the an explanation for the model overall. 

Explainers come in three types. Algorithm explainers implement a specific explanation algorithm. Generic explainers generate an explanation of a broader type, selecting the algorithm to use based on the current state-of-the-art. Base explainers are abstract classes that handle overhead code that is shared by all explainers that generate explanations of a common type. For example, the \ttt{FeatureImportance} generic explainer class extends the abstract \ttt{FeatureImportanceBase} base explainer class and wraps a \ttt{ShapFeatureImportance} or \ttt{PermutationFeatureImportance} algorithm explainer (which implement the SHAP \cite{lundberg_unified_2017} or Permutation Feature Importance \cite{breiman_random_2001} algorithms, respectively). All base explainers extend the \ttt{BaseExplainer} parent class, which contains all general-use explainer functionality such as transforming data between feature spaces, running model predictions, and formatting.

%\textbf{Explainer} objects generate explanations, and come in three types, as summarized in Figure \ref{fig:explainer_heirarchy}. 

This structure offers two benefits. First, \Pyreal{} remains agnostic to specific algorithms in favor of a focus on outputs.
%(i.e. people interacting with \Pyreal{} do not need to consider specific algorithm details, with the exception of contributors who implement the algorithms). 
Second, adding new explanation algorithms requires minimal effort and new code, because as much shared code as possible is handled by parent classes.

%\subsubsection{Transformers} \label{sec:transformers}
\textbf{Transformers} handle transforming data and explanations between feature spaces as needed to make model predictions, run explanation algorithms, and generate interpretable explanations. They have methods for three types of transforms:
%, though not all will be possible, needed, or desirable for all transformers:
transforming data from a first feature space to a second (data transform), transforming explanations from the second feature space back to the first (inverse explanation transform), and transforming explanations from the first feature space to the second (explanation transform). Note that these are the three types of transforms used by the \Pyreal{} pipeline. 

For example, the built-in \ttt{OneHotEncoder} class can one-hot encode data (data transform) and ``undo'' the encoding on explanations, resulting in the original categorical features in the explanations (inverse explanation transform). It would not be desirable to transform an explanation already presented in terms of a categorical feature to the one-hot-encoded equivalent as this would be neither more interpretable nor a better representation of the model, so the explanation transform function is not implemented.

As we have just seen, in some cases, there may not be a corresponding explanation transform or inverse explanation transform for a given data transform. In the OneHotEncoder example, the explanation transform was not desirable. In other cases, one of these transforms may be impossible; for example, a transformer that aggregates features may not be able to reverse this aggregation for feature contribution explanations. In still other cases, one of these transforms may not be needed; for example, scaling data has no impact on relative feature contribution explanations. \Pyreal{} supports missing transforms either by stopping the transform process when a missing transform is encountered (necessary if the inability to run a transform will lead to a feature space incompatible with later transforms), or by skipping it. \Pyreal{} handles all overhead code required to ensure that the explanation is presented with the same features as the corresponding data for local explanations.

The question of which transformers to use at each stage of the Pyreal pipeline is determined by three feature-space Boolean flags held by each transformer: \ttt{model}, \ttt{algo}, and \ttt{interpret}. Transformers flagged with \ttt{model} are required by the model to make predictions. Transformers flagged with \ttt{algo} are required by the explanation algorithm. Explanation algorithms usually expect the model space, so \ttt{algo} is set to \ttt{model} by default. Transformers flagged \ttt{interpret} make the data or explanations more interpretable. Figure \ref{fig:pyreal_dev_code} (lines 5-10) shows an example of these flags in action. 

\textbf{Explanation types} unify outputs from explainers that represent similar kinds of information. The other components of \Pyreal{} --- such as transformers and visualization methods --- use these objects to allow abstracted implementations. This unification reduces the number of times contributors have to write new functionality such as [inverse] explanation transform functions.

Explanation types are hierarchical. For example, local feature contributions and global feature importance are both subtypes of the \textit{feature-based} explanation type which encapsulates any explanation that gives importance scores to features. Additive local feature contributions or global feature importance further extend these types with the additional property of being able to have their importance scores added together meaningfully. Counterfactuals \cite{wachter_counterfactual_2017} and similar examples are both subtypes of the \textit{example-based} explanation type that explains model predictions with illustrative examples.

The hierarchical structure of explanation types allows for reuse of functionality between related explanation types. For instance, for any example-based explanation, the explanation is a valid row of input data to the model, so the transform-explanation function is the same as the data-transform function
Or, for any feature-based explanation that gives importance scores to features, the same bar-plotting functionality can be used

\subsection{Contributor Workflow}
Contributors add new explainers, transformers, and explanation types to the \Pyreal{} system.

Adding an explainer requires implementing the given explanation algorithm, and possibly implementing a new explanation type if the algorithm does not fit into an existing type. All other overhead work is handled by parent classes. 

Adding a transformer requires implementing the data transform function, as well as explanation transform and inverse explanation transform functions for each explanation type as needed (in practice, many transformer---explanation-type pairings do not require explanation transforms or require only very basic transforms, as seen in the examples in Table \ref{tab:sample-transforms}). In many cases, only a single [inverse] explanation transform per transformer will be required for an entire hierarchy of explanation types. 

Adding a new explanation type requires deciding on a flexible data structure to represent the explanation internally, implementing a \ttt{.validate()} function to ensure wrapped explanations meet any requirements, and possibly implementing general helper functions.

\section{Evaluation} \label{sec:evaluation}

We evaluated the interpretability of \Pyreal{} explanations, the ease-of-use of the system itself, the system's runtime performance, and the system's real-world usefulness through a series of user studies, experiments, and case studies.

Throughout our evaluations, we used ML models trained on three datasets: the California Housing dataset as introduced earlier; the Ames Housing dataset \cite{de_cock_ames_2011}, a more complex housing-price dataset with 79 input features; and the Student Performance Dataset \cite{cortez_using_2008}, which has 30 features, with each row describing a student and a target variable of whether the student will pass or fail their Portuguese class.

\subsection{Evaluating the Interpretability of Pyreal Explanations} \label{sec:explanation-interpretability}

We begin by comparing the interpretability of \Pyreal{}-generated explanations to those directly generated by the wrapped libraries that \Pyreal{} uses under-the-hood. 

\subsubsection{Participants} We conducted this study with two user groups: general participants and participants who have worked actively in the real-estate industry (who we call ``domain experts''). We further asked participants to rate the extent of their experience working with data science, statistics, ML, or AI on a scale from 0 (none) to 5 (expert). We label users who rated their experience at 2 or less as \textit{low-data-expertise} participants. 
%This group matches our intended receiver profile, and are therefore the focus of our analyses. 
Those who rated their experience as 3 or more are labelled \textit{high-data-expertise} participants.

In total, we recruited 94 participants from a random global pool using Prolific\footnote{www.prolific.com}. Table \ref{tab:participants} summarizes the distribution of these participants between the two expertise categories described above.

% \usepackage{booktabs}
% \usepackage{multirow}
\begin{table}[t]
\centering
\caption{Number of participants of different kinds of expertise in the explanation interpretability study.}
\label{tab:participants}
\begin{tabular}{@{}llcc@{}}
  \textbf{}            & \textbf{}                 & \multicolumn{2}{c}{\textbf{Data expertise (out of 5)}} \\
                     &                           & Low (0-2)           & High (3-5)          \\ \cmidrule(l){3-4} 
\multicolumn{1}{c}{\multirow{2}{*}{\textbf{\begin{tabular}[c]{@{}c@{}}Real   estate\\      expertise\end{tabular}}}} & \multicolumn{1}{c|}{No} & 28 & 36 \\
\multicolumn{1}{c}{} & \multicolumn{1}{c|}{Yes} & 17                  & 13                 
\end{tabular}
\end{table}

\begin{figure}[t]
     \centering
     \begin{subfigure}[b]{0.4\textwidth}
         \centering
         \includegraphics[width=\textwidth]{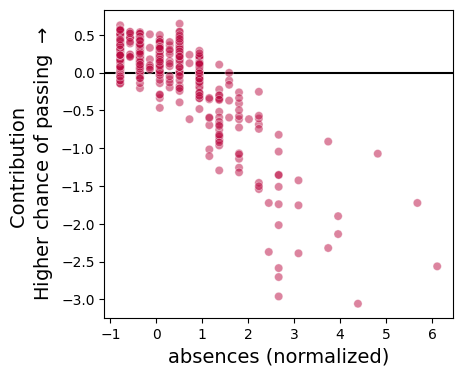}
         \caption{No Pyreal}
     \end{subfigure}
     
     \begin{subfigure}[b]{0.4\textwidth}
         \centering
         \includegraphics[width=\textwidth]{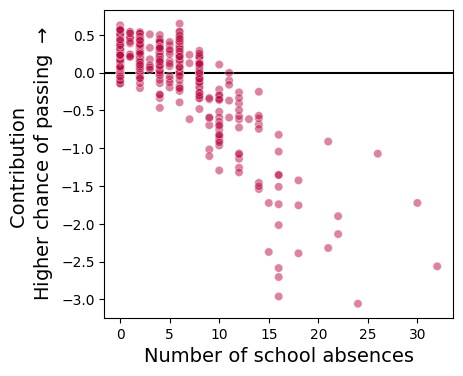}
         \caption{Pyreal}
     \end{subfigure}
        \caption{Sample figures shown to users in the explanation-interpretability user study. Participants were shown either the Pyreal explanation or the no-Pyreal explanation for each explanation.}
        \label{fig:explanation_study_figures}
\end{figure}

\subsubsection{Method} Participants were shown explanations of model predictions on the California Housing dataset and the Student Performance dataset. The explanations were feature contribution explanations generated using the SHAP algorithm, either in the form of a bar plot (explaining a single prediction, as shown in Figure \ref{fig:motivation}) or a scatter plot (showing contributions across the dataset for a single feature, as shown in Figure \ref{fig:explanation_study_figures}). Each explanation was either in the interpretable feature space (\taskpyrealclean{} condition) or in the model-ready feature space (\tasknopyrealclean{} condition). 

In total, we had 2 datasets \texttimes{} 2 conditions \texttimes{} 5 individual plots (3 bar, 2 scatter) = 20 explanations. Each participant was shown one condition for one dataset, and the other condition for the other dataset (10 total explanations per participant). The within-subjects design allowed us to account for differences between participants, while the between-datasets design prevented participants from learning about the meaning of features from \Pyreal{} explanations or vice versa. 

For each explanation, participants were asked to 1) rate the usefulness of the explanation on a Likert-type scale from 1 (not at all useful) to 5 (extremely useful), 
%2) describe what the explanation told them about the model's logic, 
2) describe what made the explanation more useful, and 3) describe what made the explanation less useful. Answers 2 and 3 were given in open response format.

\subsubsection{Results (general participants)} 
We begin by analyzing the results from the general participant (no real estate expertise) group, with extra consideration for the low-data-expertise group that represents our target receiver audience. 

\begin{figure}[tb]
     \centering
     \begin{subfigure}[b]{.78\linewidth}
         \centering
         \includegraphics[width=\linewidth]{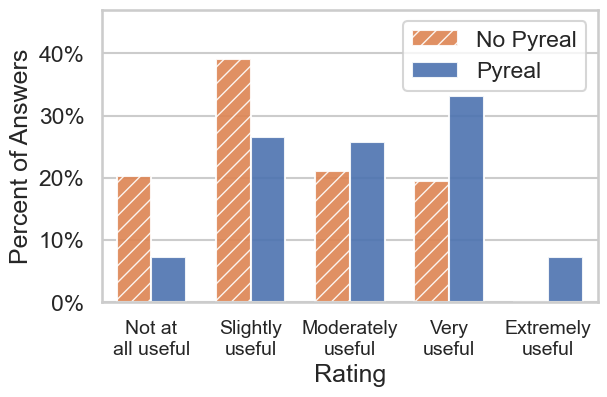}
         \caption{Usefulness scores given by low-data and no-domain expertise participants.}
         \label{fig:usefulness-low-data}
     \end{subfigure}

     \begin{subfigure}[b]{.78\linewidth}
         \centering
         \includegraphics[width=\linewidth]{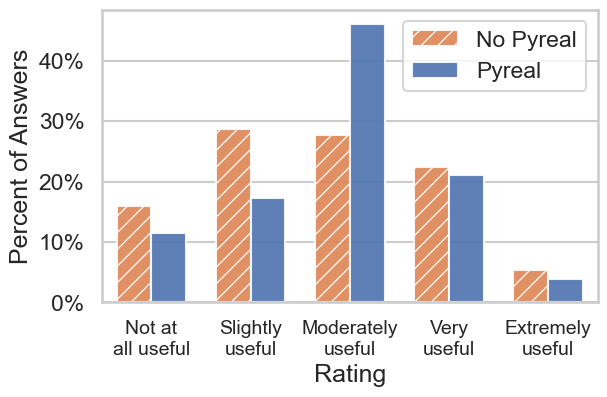}
         \caption{Usefulness scores given by high-domain-expertise participants on the real estate dataset explanations.}
         \label{fig:usefulness-real-estate}
     \end{subfigure}
        \caption{Likert scale responses to the question ``how useful did you find the explanation?'' in our explanation-interpretability evaluation from two key participant groups representing target receivers.}
        \label{fig:usefulness}
\end{figure}

The average usefulness scores (out of 5) given by the low-data-expertise participants on each condition were 3.06 (Pyreal) versus 2.40 (No-Pyreal). Low-data-expertise participants rated Pyreal explanations as being extremely useful and very useful for 7.23\% and 33.06\% of explanations respectively, compared to 0\% and 19.51\% of explanations for the no-Pyreal explanations. 
Figure \ref{fig:usefulness-low-data} shows the distribution. As expected given \Pyreal{}'s target receiver audience, the high-data-expertise participants saw less usefulness from \Pyreal{}, with average usefulness ratings of 2.73 (Pyreal) versus 3.01 (No-Pyreal).

We conducted a thematic analysis on the open-response answers to further investigate what made explanations more or less useful to low-data-expertise participants. Here, we describe the common themes found for open-response questions on both Pyreal and non-Pyreal explanations.

\textbf{Interpretable features are essential for understanding.} One common theme we identified for non-Pyreal explanations was confusion related to the model-ready feature space used, which included normalized feature values. Of the 23 participants who answered the question ``what made the explanation less useful'' on any no-Pyreal explanation, 21 reported difficulties understanding the information (features) presented or difficulties interpreting the normalized feature values. For example, P22 said, ``\textit{what does -0.58 bedrooms mean? No range, no context}''. P14 said, ``\textit{what does `fjob\_other' mean?}'' and ``\textit{longitude and latitude separately is strange}'' (the equivalent Pyreal explanations showed location in terms of neighborhood instead of lat/long values). 

In comparison, only 9 participants reported any confusion related to understanding features on the Pyreal explanations, and most of these concerns related to wanting more detailed information about how the features were computed or what they meant in context. For example, P7 reported that they did not understand the relationship of the characteristics (features) to the purpose of the model.
%, and requested a glossary of features used and their details. 
%We believe these kinds of concerns would be less of an issue in a real-world use case where explanation receivers are domain experts, or would otherwise receive further context on the problem beyond the brief introduction we were able to give for this study. 
Only one comment suggested that using the interpretable feature space reduced the usefulness of the explanation in any way, commenting that the Pyreal explanation (showing the neighborhood rather than lat/long values) did not provide an ``accurate location''. 

These findings motivate our system of explanation transforms to ensure that receivers are given explanations in an interpretable feature space.
%This is why we need explanation transforms and find their equivalent in a intepretable feature space. 

%Four participants were not able to identify anything useful about non-Pyreal explanations, even when prompted. This was not the case for any participants on Pyreal explanations. 

\textbf{Participants did not shy away from calling the no-Pyreal explanations ``useless"} Another common theme seen only in comments on no-Pyreal explanations were direct and unspecific complaints about the explanations' lack of clarity. P6 commented on a no-Pyreal explanation ``\textit{this is a very useless method of data visualization}''. P3 said on another no-Pyreal explanation ``\textit{the factors are very vague and not very helpful}''. P13 described the non-Pyreal explanations as ``\textit{totally not relatable}'', and P4 said simply ``\textit{everything [makes this explanation less useful], I do not understand this}''.

\textbf{With Pyreal, participants were able to dig deeper into the ML model itself.} A theme in comments on Pyreal explanations was concerns about the model's prediction logic itself (7 cases for Pyreal explanations, but only 1 for non-Pyreal). For example, P14 said with regards to a Pyreal explanation ``\textit{Median income has too great impact on price prediction}''. The goal of ML explanations is to enable participants to understand where ML model predictions come from, and because the model logic was the same in both conditions, this finding suggests that Pyreal explanations were more successful.

%We  note that despite instructions to the contrary, some participants did report reducing their usefulness rating of explanations due to disagreeing with the model's logic; we believe factoring this in could increase the average rating of Pyreal's usefulness.

\subsubsection{Results (domain expert participants)}
We now consider the results from domain expert participants (those who reported working in the real estate industry). For these participants, we only consider responses to explanations on the California Housing dataset, on which their domain expertise applied.

The reported usefulness of \Pyreal{} was slightly lower among domain experts, with average usefulness scores of 2.88 (Pyreal) versus 2.72 (no-Pyreal) (there was no significant difference in these average scores between domain experts with low or high data experience). Figure \ref{fig:usefulness-real-estate} shows the distribution. This finding was not surprising, as it is expected that people have higher standards for information about their field. Additionally, as discussed below, domain experts may require more detail about the context around decision-making problems in their area. 

\textbf{Domain expert participants felt no-Pyreal explanations were not appropriate for them.} Several participants mentioned that they felt they lacked the necessary knowledge to understand the no-Pyreal explanations, despite being experts in the domain. P87 said about a no-Pyreal explanation: ``\textit{Not understandable values for mathematical non-literate people.}''. P89 said, ``\textit{I'll be honest that I'm not a numbers guy and I don't understand what `normalised' means.}'' This theme did not appear in comments on \Pyreal{} explanations.

\textbf{The decision-making context is especially important to domain experts, and carefully crafted features can help establish this.} One source of confusion we saw from domain experts with the Pyreal explanations on the real estate dataset related to misunderstanding or needing more details about the broader context. For example, in the California Housing Dataset, each row of data refers to a \textit{block of houses}, but most people in the domain tend to be used to reasoning about the prices of \textit{individual houses}. The study instructions gave this context, but the scenario may have been so far out of the norm for participants that they regardless expected features to be presented in terms of individual houses. For example, P88 incorrectly assumed that ``\textit{...`number of households' refers to the number of households that would occupy a single home}'' and P95 reported being unsure why the total number of bedrooms was 190 (a value too high for a single house). 
This finding reveals two lessons: first, ML test datasets like this one often do not effectively represent real-world decision-making problems, which may affect the effectiveness of user studies. Second, this further reveals the impact that the format of features may have on user understanding.

\textbf{Domain expert participants had similar needs for interpretable features.} Like in our general populace study, the domain experts frequently cited a lack of understanding about what the features or their normalized values referred to. 13 out of 19 participants who commented on no-Pyreal graphs made at least one comment about not understanding the features or their values. For example, P91 said, ``\textit{...what does -0.62 mean for bedrooms? That the home has less than one room?}'' For comparison, 5 out of the 10 participants who commented on Pyreal graphs expressed confusion related to not understanding features, though 3 of these 5 appeared to be confused about the context as explained above (thinking rows referred to individual houses instead of blocks, and thus being confused about large feature values). 

\subsubsection{Further Investigation} To further investigate the usefulness of Pyreal explanations, we conducted a second study that enabled participants to more directly compare conditions. For this study, we recruited 20 new participants, 12 of whom rated their ML expertise as 2 or less out of 5.

For this study, we presented the same set of explanations, but showed two explanations at a time side by side. One was generated using Pyreal in the interpretable feature space, and the other was the equivalent explanation generated directly with wrapped libraries in the model feature space. 
%Both explanations showed the same information but in a different feature space.The explanations used were the same as in the previous study. 
Each participant was shown 10 pairings.

For each pair of explanations, we asked users to select the one they thought was more useful. Among the 12 low-data-expertise participants, the Pyreal explanation was selected as more useful in 110 out of 120 total pairings  (91.67\%). Among high-data-expertise, the Pyreal explanation was selected as more useful in 61 out of 80 total pairings (75.25\%).

\begin{figure}[tb]
    \centering
    \begin{lstlisting}[language=Python]
   from sklearn.preprocessing import OneHotEncoder
   import shap
   import reverse_geocoder
   
   X_train = load_training_data()  # external function
   X_to_explain = load_data_to_be_explained()

   def transform_to_x_model(X):
       x_to_encode = X[["ocean_proximity"]]
       ohe = OneHotEncoder().fit(x_to_encode)
       encoded_columns = ohe.get_feature_names_out()
       ocean_encoded = ohe.transform(x_to_encode)
       return pd.concat([
            X.drop("ocean_proximity", axis="columns"), 
            ocean_encoded], axis=1)

   X_train_model = transform_to_x_model(X_train)
   X_model = transform_to_x_model(X_to_explain)
   
   X_interpret = X_to_explain.copy()
   coordinates = list(
        X_interpret[["latitude", "longitude"]]
        .itertuples(index=False, name=None))
   results = reverse_geocoder.search(coordinates)
   neighborhoods = [result["name"] for result in results]
   X_interpret["neighborhood"] = neighborhoods
   X_interpret = X_interpret.drop(
        columns=["latitude", "longitude"])
        
   model = train_model(X_train_model)  # external function
   explainer = shap.Explainer(model, X_train_model)
   shap_exp = explainer(X_model)
   columns = X_model.columns
   exp_df = pd.DataFrame(shap_exp.values, columns=columns)
   encoded_features = 
        [item for item in encoded_columns 
        if item.startswith("ocean_proximity_")]
   exp_df = exp_df.drop(encoded_features, axis="columns")
   exp_df["ocean_proximity"] = 
        exp_df[encoded_features].sum(axis=1)
   exp_df["neighborhood"] = 
        exp_df["latitude"] + exp_df["longitude"]
   exp_df = exp_df.drop(columns=["latitude", "longitude"])

   feature_descs = {"med_income": "Median income", ...}
   exp1 = exp_df.rename(feature_descs, axis="columns") \end{lstlisting}
  \caption{The equivalent code required to generate the same feature contributions explanation (\ttt{exp1}) generated by the code seen in Figures \ref{fig:pyreal_user_code} and \ref{fig:pyreal_dev_code}. Note that this code only generates one kind of explanation. To generate another, lines 32-50 would need to be modified for the new explanation, accomplishing the equivalent of line 8 in Figure \ref{fig:pyreal_user_code}.} \label{fig:pyreal_equivalent_code}
\end{figure}

\subsection{Evaluating the Usability of the Pyreal System}
We now evaluate the ease-of-use of the \Pyreal{} system and the simplicity of its code. To begin with, for context, we offer a demonstration of the equivalent code required to generate an interpretable explanation without \Pyreal{}, shown in Figure \ref{fig:pyreal_equivalent_code}. Comparing this code to Figures \ref{fig:pyreal_user_code} and \ref{fig:pyreal_dev_code} illustrates the relative simplicity and readability of \Pyreal{} code compared to alternatives. Additionally, one can see the benefit of \Pyreal{}'s explanation-switching API through comparing the additional effort required to add an explanation type.

Next, to formally evaluate how easy it is to generate interpretable explanations with \Pyreal{}, we discuss our usability study in which we asked participants to use Pyreal to generate explanations.

\subsubsection{Participants} For this study, we had seven participants. Two were ML experts, and five had some ML experience.

\subsubsection{Method} 
%This study was run using a Google Colab notebook. 
Participants were first asked to go through a brief tutorial on developing Pyreal applications, using a synthetic trinket pricing dataset. They then began the real task, which used the California Housing dataset. Participants were told the performance of the house price prediction model, including the mean-absolute-error on house price (\$29,823), the $r^2$ score (0.80), and a scatter plot of real house price to predicted house price. Participants were then asked to generate a local feature contribution explanation using \Pyreal{}
%and answer questions about what contributed to the model's prediction on a sample housing block. The purpose of these questions was to ensure participants were thinking about

Finally, participants were asked to rate \Pyreal{} using a five-point Likert scale on several axes: the usefulness of \Pyreal{} explanations, how easy \Pyreal{} was to learn, and how likely they were to use \Pyreal{} if a need for explanations arose. 

\subsubsection{Results} On average, participants rated \Pyreal{} as better-than-neutral (out of 5) on all metrics for the system's usefulness ($\mu=3.14, \sigma=.90$), ease of learning ($\mu=3.42, \sigma=0.98$), and likeliness of use ($\mu=3.71, \sigma=0.49$). In terms of system usability only one participant disagreed that \Pyreal{} was easy to learn. Five participants agreed they would be likely to use \Pyreal{} in the future if the opportunity arose, with the other two participants being neutral on this question.

%\subsection{Evaluating the Complexity of Pyreal Code}

% \usepackage{booktabs}
% \usepackage{graphicx}
\begin{table}[]
\caption{Summary of the conditions considered in our evaluations.}
\label{tab:study_conditions}
\resizebox{\columnwidth}{!}{%
\begin{tabular}{@{}lll@{}}
\toprule
                   & \textbf{Libraries Used} & \textbf{Output Format}      \\ \midrule
\tasknopyreal      & SHAP/sklearn            & Model-ready explanation \\[3pt]
\tasknopyrealinter & SHAP/sklearn            & Interpretable explanation   \\[3pt]
\taskpyreal        & Pyreal                  & Interpretable explanation   \\[3pt]
\taskpyrealformat & Pyreal & \begin{tabular}[c]{@{}l@{}}Interpretable explanation in\\ usable data structure format\end{tabular} \\ \bottomrule
\end{tabular}%
}
\end{table}

\subsection{Evaluating the Runtime Performance of Explanation Generation}
We evaluated the runtime impacts of Pyreal and interpretable explanation generation, to determine 1) the extent to which generating interpretable explanations is slower than the default, model-space explanations and 2) whether Pyreal has any inefficiencies that cause a slower runtime than would occur from generating the same explanations without Pyreal. 
%We consider both the theoretical runtime impacts and the observed impacts based on experiments.

\subsubsection{Method} In these evaluations, we consider the runtime of four tasks, summarized in Table \ref{tab:study_conditions}. We generate interpretable explanations directly using wrapped libraries (\tasknopyrealinter{}), using Pyreal (\taskpyreal{}), and using Pyreal with additional output formatting for usability (\taskpyrealformat{}). We also generate model-space explanations, the default explanations from wrapped libraries (\tasknopyreal{}).

%1) generating model-space explanations, the default explanations from wrapped libraries (\ttt{task:model}), 2) generating interpretable explanations directly using wrapped libraries (\ttt{task:no\_pyreal}), 3) generating interpretable explanations using Pyreal (ie. the output of explainer objects, in Pyreal Explanation formats) (\ttt{task:pyreal}), and 4) generating interpretable explanations with additional output formatting for usability (ie. the output of RealApp objects, in a usable format such as a dictionary of DataFrames) (\ttt{task:realapp}). For all tasks, ``generating an explanation'' includes fitting the explainer and producing explanation, along with any data/explanation transforms need to do so. 
%Fitting the ML model and transformers is not included.

\begin{figure}[tb]
    \centering
    \includegraphics[width=.9\linewidth]{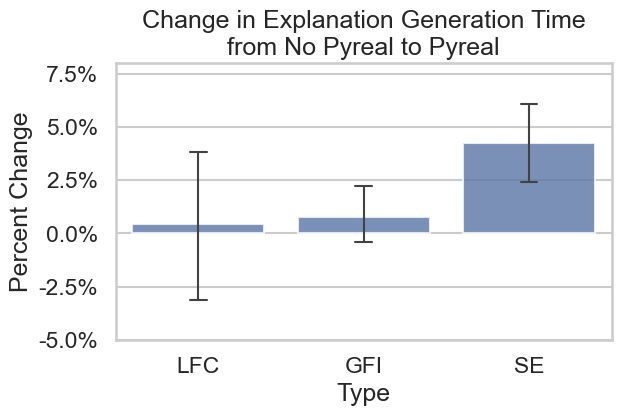}
    \caption{Average percent change in explanation generation time for generating an interpretable explanation with \Pyreal{} (\taskpyreal{}) compared to generating the same explanation without \Pyreal{} (\tasknopyrealinter{}). \Pyreal{} incurred at most an 8.64\% increase in runtime.  
    LFC: local feature contributions. GFI: global feature importance. SE: similar examples.}
    \label{fig:runtime}
\end{figure}

Our evaluation used three explanation types: local feature contributions (LFC) using Tree SHAP or Linear SHAP depending on the dataset, global feature importance (GFI) using \ttt{sklearn}'s permutation feature importance, and similar examples (SE) using \ttt{sklearn}'s \ttt{KDTree} module \cite{pedregosa_scikit-learn_2011}
%The latter is an example-based local explanation that gives rows from the training set that are similar to the input row and the model prediction on those rows.

We used all three evaluation datasets and made 10,000 and 20,000 row versions of each. In other words, we generated explanations on a 10,000 row testing dataset using a 10,000 row training dataset (both datasets were randomly sampled from the original dataset). We then repeated the process with 20,000 row testing and training datasets\footnote{For permutation feature importance on the Ames Housing dataset, we used smaller datasets of 1,000 and 2,000 rows, as this algorithm is very slow on datasets with many columns}.

%All tasks, explanation types, datasets, and data sizes resulted in 72 experiments. 
For each experiment (task/explanation-type/dataset/data size), we produced the explanation 10 times and computed the average runtime.

\subsubsection{Results} 
%Figure \ref{fig:runtime} summarizes the observed change in explanation generation time introduced by making an explanation interpretable or using \Pyreal{}.
%in terms of explanation generation time relative to a baseline of the model-space explanation generation task (\ttt{task:model}).
The increase in runtime when generating an interpretable explanation compared to a model-space explanation varies depending on the transforms used and the runtime of the base explanation algorithm. Our highest percent change from \tasknopyreal{} to \taskpyreal{} was 86.86\% on the Ames Housing dataset (10,000 rows) for LFC. Our explanation transforms in this case ran in linear time, while the base explanation algorithm (Linear SHAP) runs in constant time, resulting in a relatively higher percent increase in runtime. In contrast, we saw almost no change in runtime for the GFI tasks. This was expected, as the permutation feature importance algorithm's runtime scales with respect to both number of features and number of rows, while explanation transforms on GFI explanations have a constant runtime, as these explanations include only one row of data. We saw a slight increase in runtime for SE explanations, as we needed to transform the dataset to the interpretable space to return interpretable examples.

Pyreal generates interpretable explanations in a similar amount of time as non-Pyreal approaches, with the percent change in generation time between \tasknopyrealinter{} and \taskpyreal{} ranging from $-7.74\%$ to $8.64\%$. Figure \ref{fig:runtime} shows this result.

For GFI and SE there was little increase in runtime from formatting. For LFC, formatting introduced a linear-time process (converting a DataFrame to a dictionary of DataFrames per row being explained) to a constant-time explanation algorithm (LinearSHAP), so we saw higher slowdowns of up to 106\% per 1,000 rows of data. This formatting will generally not be run on more data than a human can process, so real time increases should be small (up to about .4 seconds per 1,000 rows on a AMD 5800x processor).

\subsection{Evaluating Pyreal in the Real-World: Case Studies}
We have applied \Pyreal{} to two real-world domains: child welfare screening \cite{zytek_sibyl_2021} and wind turbine monitoring \cite{zytek_lessons_2023}. In both domains, we identified a need for interpretable features in the explanations shown to receivers. 
For child welfare screening, we used Pyreal to generate human-readable local feature contribution explanations that could be understood and readily used by child welfare screeners who are deciding which cases of potential child abuse to investigate. These screeners were not familiar with ML, so the default one-hot-encoded and confusingly named features in explanations were not useful. Pyreal allowed us to present readable explanations using categorical feature names, as well as present Boolean features using positive and negative language (ex. \ttt{The child does not have a sibling}, rather than \ttt{The child has a sibling --- False}). 

For wind turbine monitoring, collaborators were unsure of which types of explanations would best aid with decision-making. \Pyreal{}'s low-code high-level API enabled us to easily experiment with multiple explanation types, all presented in the format most natural to users, to tune our explanation interface for the specific needs of the context.

\section{Discussion} \label{sec:discussion}

Our studies suggest that \Pyreal{} creates \interpretable{} explanations that are perceived by our target receiver audience to be more useful than uninterpretable equivalents, and result in less confusion. Answers to free-response questions suggested that participants are more able to identify potential correct and incorrect elements of the model's logic. We have also confirmed that \Pyreal{} does not impose a significant runtime penalty over existing solutions. 
%when generating these interpretable explanations.

Our usability study and real-world case studies suggest that \Pyreal{} is easy to use and quick to learn, though we would like to investigate this further. Future studies will empirically examine the system's usability with a larger audience within a real-world decision-making problem.

While in this paper we focused on the systems questions posed by explanation transforms, future work should investigate the theoretical side of the problem through a formal verification of the fidelity of transformed explanations.

Additionally, in this paper we have taken one essential step towards highly usable ML explanations by offering interpretability. Another important property of explanations is usefulness, or offering the correct types of explanations for a given decision-making problem. This is a context-dependant task, and at the time of writing, we are actively working on live deployments in real-world domains to tackle this other side of the usable ML problem.

We are continuing to develop \Pyreal{}'s library implementation, adding additional explanation algorithms, types, and transformers. Each added component increases the scope of domains where \Pyreal{} can be useful, opening up new domains for real-world use-cases and evaluations. Future work may also expand \Pyreal{} to other data modalities such as image and text.

\section{Conclusion} \label{sec:conclusion}
We developed \Pyreal{}, a highly-extensible system for generating interpretable ML explanations. \Pyreal{} handles the data and explanation transform pipeline required to generate such explanations under the hood. This allows even users without extensive experience with explainable ML to generate interpretable and valuable explanations for use in decision-making. 

\begin{acks}
    We would like to thank Arash Akhgari for work on our graphics and visualizations, and Cara Giaimo for feedback on writing. We would like to thank Asher Norland for development work on the \Pyreal{} library. Finally, we would like to thank our anonymous reviewers for their insights and feedback.
\end{acks}

\bibliographystyle{ACM-Reference-Format}
\interlinepenalty=10000
\bibliography{references}

\appendix

\end{document}